\documentclass{llncs}

\usepackage{url}
\usepackage{graphics}
\usepackage{tabularx}
\usepackage{multirow}
\usepackage{rotating}
\usepackage{alltt}
\usepackage{amsmath}
\usepackage{amssymb}
\usepackage{amsfonts}
\usepackage{longtable}
\usepackage{pdflscape}
\usepackage{verbatim}
\newcommand\R{\mathbb{R}}

\begin{document}

\mainmatter

\title{XES Tensorflow -- Process Prediction using the Tensorflow Deep-Learning Framework}
\subtitle{Demo Paper}

\titlerunning{XES Tensorflow}

\author{Joerg Evermann\inst{1} \and Jana-Rebecca Rehse\inst{2,3} \and Peter Fettke\inst{2,3}}

\authorrunning{Joerg Evermann et al.}

\institute{Memorial University of Newfoundland
\and
German Research Center for Artificial Intelligence
\and
Saarland University}

\maketitle

\begin{abstract}
Predicting the next activity of a running process is an important aspect of process management. Recently, artificial neural networks, so called deep-learning approaches, have been proposed to address this challenge. This demo paper describes a software application that applies the Tensorflow deep-learning framework to process prediction. The software application reads industry-standard XES files for training and presents the user with an easy-to-use graphical user interface for both training and prediction. The system provides several improvements over earlier work. This demo paper focuses on the software implementation and describes the architecture and user interface. 


\end{abstract}

\section{Introduction}

The prediction of the future development of a running case, given information about past cases and the current state of the case, i.e. predicting trace suffixes from trace prefixes, is a core problem in business process intelligence (BPI). Recently, deep learning \cite{LeCunetal:2015,DBLP:journals/nn/Schmidhuber15} with artificial neural networks has become an import predictive method due to innovations in neural network architectures, the availability of high-performance GPU and cluster-based systems, and the open-sourcing of of multiple software frameworks at a high level of abstraction.

Because of the sequential nature of process traces, recurrent neural networks are a natural fit to the problem of process prediction. An initial application of RNN to BPI \cite{Evermannetal:2016} used the executing activity of each trace event as both predictor and predicted variable. Evaluation on the BPI 2012 and 2013 challenge datasets showed training performance of up to 90\%, but that study did not perform cross-validation. A more systematic study \cite{EvermannRF2017}, including cross validation, showed significant overfitting of the earlier results, i.e. the neural network capitalizes on idiosyncrasies of the training sample that do not generalize. The later work also showed that the size of the neural network has a significant effect on predictive performance. Further, predictive performance can be improved when organizational data is included as predictor. Both \cite{EvermannRF2017} and \cite{Evermannetal:2016} use approaches in which each ``word'' (e.g. the name of the executing activity of each event) is embedded in a k-dimensional vector space (details in Sec.~\ref{sec:inputs} below), which forms the input to the neural network. In contrast, an independent, parallel effort \cite{DBLP:journals/corr/TaxVRD16} eschews the use of embeddings, encoding event information as numbered categories in one-hot form. That approach also adds time-of-day and day-of-week as additional predictors. Both approaches compare the predicted suffix to the target suffix by means of a string edit distance, showing similar performance. Additionally, \cite{EvermannRF2017} demonstrates that predicted suffixes are similar to targets by showing similar replay fitness on a model mined from the target traces.

The software application described here extends the previous approaches in a number of ways. Primarily, it is more general and flexible with respect to the case and event attributes that can be used as predictor or predicted variables. Whereas \cite{EvermannRF2017,DBLP:journals/corr/TaxVRD16} use only the activity name and lifecycle transition of an event, and \cite{EvermannRF2017} also includes resource information of an event as predictor, our application can use any case- or event-level attribute as predictor. We can also predict any and multiple event attributes of subsequent events. These advantages are due to differences in input encoding. Whereas \cite{EvermannRF2017} concatenates activity name, lifecycle transition and resource information into an input string, which is then assigned a category number and embedded in a vector space for input to the neural network, our approach assigns category numbers to each input attribute separately, then embeds them into their own vector spaces and concatenates the embedding vectors to form the input vector. Details are presented in Sec.~\ref{sec:inputs}. The advantage is much smaller input ``vocabularies'' (sets of unique inputs) which can be adequately represented by much smaller input vectors. It also allows easy mixing of categorical predictors with embedding inputs and numerical predictors which are directly passed to the neural net, simply by concatenating these inputs. 

Additionally, our approach can predict multiple variables, for example the activity as well as the resource information of the next event. We use either shared or separate RNN layers for each predicted attribute. 

Finally, the software tool presented in this demo paper includes a graphical user interface that guides novice users and avoids the need for specialized coding, it provides integration with a graphical dashboard, and a stand-alone prediction application with an easy-to-use prediction API. 

As a demo paper, this paper focuses on the software implementation, architecture and user interface with only a short exposition of the neural network background. In particular, we describe the input and output handling of the neural network (Sections~\ref{sec:inputs}, \ref{sec:outputs}), as this is where our approach differs from \cite{EvermannRF2017,DBLP:journals/corr/TaxVRD16}.  More details on neural networks in general can be found in \cite{LeCunetal:2015,DBLP:journals/nn/Schmidhuber15}, and, applied to process prediction, in \cite{EvermannRF2017,DBLP:journals/corr/TaxVRD16}. Our software is open-source and available from the authors\footnote{\url{http://joerg.evermann.ca/software.html}}.

\section{Recurrent Neural Networks Overview}

\begin{figure}
\centering
\includegraphics[width=\linewidth]{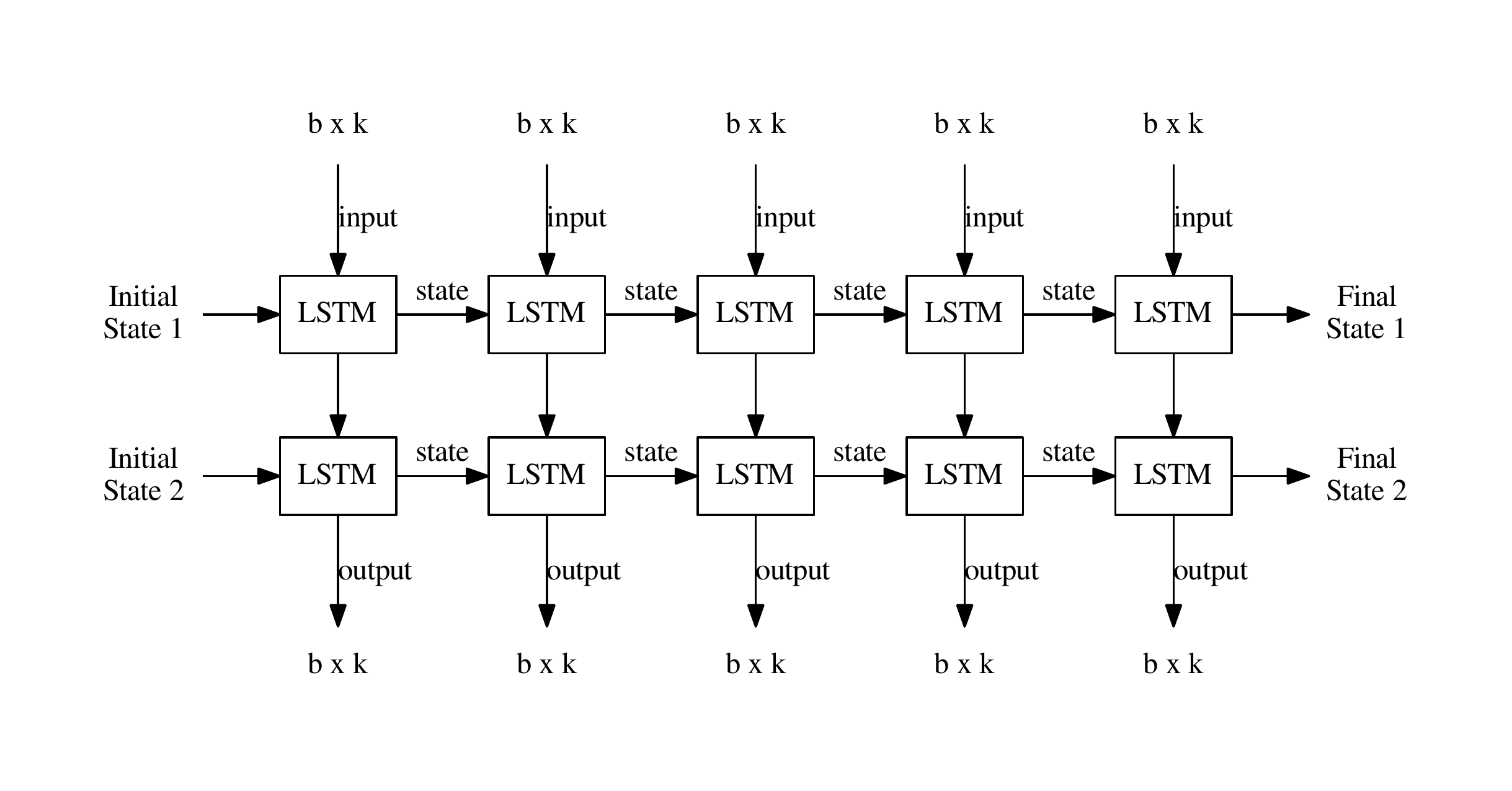}
%\vspace{-35pt}
\caption{An RNN with 5 unrolled steps, 2 layers, batch size $b$ and input length $k$}
\label{fig:rnn}
\end{figure}

A recurrent neural network (RNN) is one in which network cells can maintain state information by feeding the state output back to themselves, often using a form of long short term memory (LSTM) cells \cite{DBLP:journals/neco/HochreiterS97}. To make this feedback tractable within the context of backpropagation, network cells are copied, or ``unrolled'', for a number of steps. Fig.~\ref{fig:rnn} shows an example RNN with LSTM cells; detailed descriptions can be found in \cite{EvermannRF2017,DBLP:journals/corr/TaxVRD16}.

%\vspace{-25pt}

\section{Tensorflow}

Our software implementation is based on the open-source Tensorflow deep learning framework\footnote{\url{https://www.tensorflow.org}}. Tensors are generalizations of matrices to more than two dimensions, i.e. n-dimensional arrays. A Tensorflow application builds a computational graph using tensor operations. A loss function is a graph node that is to be minimized. It compares computed outputs to target outputs in some way, for example as cross-entropy for categorical variables, or root mean squared error for numeric variables. Tensorflow computes gradients of all tensors in the graph with respect to the loss function and provides various gradient descent optimizers. Training proceeds by iteratively feeding the Tensorflor computational graph with inputs and targets and optimizing the loss function using back-propagated errors. 

\section{Training}

Our software system consists of two separate applications, one for training a deep-learning model, and another one for predicting from a trained model. This section describes the training application. 

\subsection{XES Parser}

Training data is read from XES log files \cite{xes}, beginning with global attribute and event classifier definitions. Using these, the traces and their events are read. While the XES standard allows traces and events to omit globally declared attributes, it does not allow specification of default values for missing attributes. Hence, the XES parser omits any incomplete or empty traces. String-typed attributes are treated as categorical variables. Their unique values (categories) are numbered consecutively, encoding each as an integer in $0 \ldots l_i$, where $l_i$ is the number of unique values for attribute $i$. Datetime-typed attributes are converted to relative time differences from the previous event. The user can choose whether to standardize or to scale them to days, hours, minutes or seconds for meaningful loss functions. Numerical attributes are standardized. For multi-attribute event classifiers, the parser constructs joint attributes by concatenating the string-typed attributes or by multiplying the numerically-typed attributes specified in the classifier definition. End-of-case markers are inserted after each trace. 

\subsection{Inputs}
\label{sec:inputs}

RNN training proceeds in ``epochs''. In each epoch, the entire training dataset is used. Before each epoch, the state of the RNN is reset to zero while trained network parameter values are retained. Within each epoch, training proceeds in batches of size $b$ with averaged gradients to avoid overly large fluctuations of gradients. The batch size $b$ can be freely chosen by the user.

For each unrolled step $s$, the RNN accepts a floating point input tensor $I_s \in \R^{b \times p}$, where $b$ is the batch size and $p$ can be freely chosen. Numerical and datetime predictors are encoded directly, each yielding a floating point tensor $I_{s, i} \in \R^{b \times 1}$ for predictor attribute $i$. Categorical attributes, encoded as integer category numbers, are transformed using an embedding matrix $E_i \in \R^{l_i \times k_i}$. This is a lookup matrix of size $l_i \times k_i$, where $l_i$ is the number of categories for predictor attribute $i$ and $k_i$ can be freely chosen. Embedding lookup transforms an integer category number $j_{s,i}$ to a floating point vector of size $k_i$. The larger the value of $k_i$, the better the attribute values are separated in the $k_i$-dimensional space. At the same time, larger $k_i$ lead to increased computational demands. The output of the embedding lookup $E(.)$ is a tensor $I_{s, i} \in \R^{b \times k_i} = E(j_{s, i})$. The tensors for all predictor attributes are concatenated, yielding a tensor $C_s \in \R^{b \times m}$ where $m$ is the sum of the second dimensions of the concatenated tensors. An input projection $P^I_s \in \R^{m \times p}$ can be applied so that the input to each unrolled step of the RNN is $I_s = C_s \times P^I_s$.

\subsection{Model}

In our approach, the user can train a model on multiple predicted event attributes (``target variables'') concurrently, for example predicting the activity as well as the resource of the next event. For this, the user can choose to share the RNN layers across the different target variables, or to construct a separate RNN for each target variable. The input embeddings are shared in all cases.

\subsection{Outputs}
\label{sec:outputs}

The output of the RNN for each unrolled step is a tensor $O_{s} \in \R^{b \times p}$. For a categorical predicted variable $i$, this is multiplied by an output projection $P^O_{s, i} \in \R^{p \times l_i}$ to yield $O_{s, i} \in \R^{b \times l_i} = O_s \times P^O_{s, i}$. A softmax function is applied to generate probabilities over the $l_i$ different categories $S_{s, i} \in \R^{b \times l_i} = \text{softmax}_{l_i}(O_{s, i})$. A cross-entropy loss function $L_i = H_{S_{s, i}}(T_{s, i})$ is then applied, comparing the output probabilities against ``one-hot'' encoded targets $T$. One-hot encoding is a vector with the element indicating the target category number equal to one and the remainder set to zero. For numerical attributes, the output $O_s$ is multiplied by a $P^O_{s, i} \in \R^{p \times 1}$ output projection, yielding $O_{s, i} \in \R^{b \times 1} = O_s \times P^O_{s, i}$. This is compared to target values $T_{s, i}$ using the mean square error (MSE) $L_i = \overline{(O_{s, i} - T_{s, i})^2}$, the root mean square error (RMSE) $L_i = \sqrt{\overline{(O_{s, i} - T_{s, i})^2}}$, or the mean absolute error (MAE) loss function $L_i = \overline{|O_{s, i} - T_{s, i}|^2}$.

\subsection{Logging and TensorBoard Integration}

\begin{figure}[b]
\centering
\includegraphics[width=\textwidth]{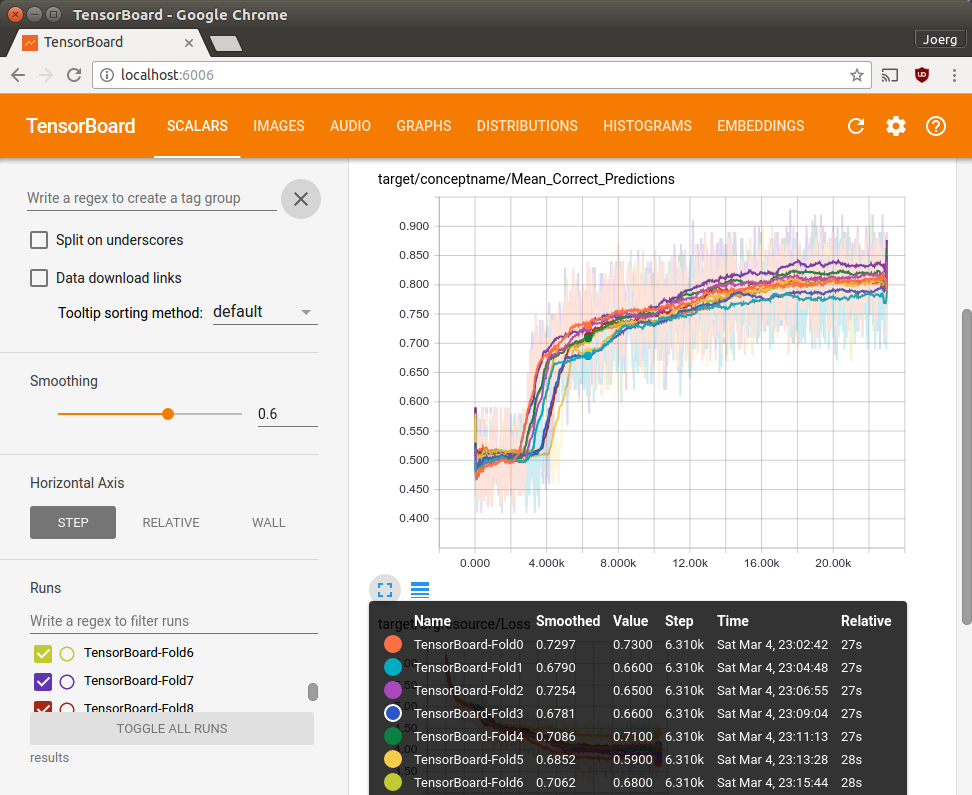}
\caption{Tensorboard dashboard showing training performance for 10 folds}
\label{fig:tensorboard}
\end{figure}

Our software logs summary information about the proportion of correct predictions and the loss function value for each training step. The embedding matrices for all categorical predictor variables are saved at the end of training, together with the computational graph. This information can be read by the Tensorboard tool (Fig.~\ref{fig:tensorboard}) to visualize the training performance, the graph structure, and to analyze the embedding matrices using t-SNE or principal components projection into two or three dimensions. Finally, the entire trained network is saved in a ``frozen'', compacted form to be loaded into the prediction application.

\section{Prediction}

The prediction application loads a trained model, saved by the training application, as well as the corresponding training configuration, and predicts trace suffixes from trace prefixes. Trace prefixes are read from XES files. Because the trained model expects batches of size $b$, at most $b$ trace prefixes are loaded at a time. The network state is initialized to zero and the trace prefixes are input to the trained network, encoded as described in Sec.~\ref{sec:inputs}. The network outputs for the last element of a trace prefix are the predictions for the attributes of the next event. For categorical attributes, the integer output indicating the category number is translated back to the character string value. For datetime-typed attributes, the attribute value is computed by adding the predicted value to the attribute value of the prior event. The predicted event is then added to the trace prefix and the prediction process can be repeated. In this way, suffixes of arbitrary length can be predicted. The user can stop prediction when an end-of-case marker has been predicted, or after a specified number of events have been predicted. The predicted suffixes are written back to an XES file. 

\section{Software Implementation}

\begin{figure}
\centering
\includegraphics[scale=0.4]{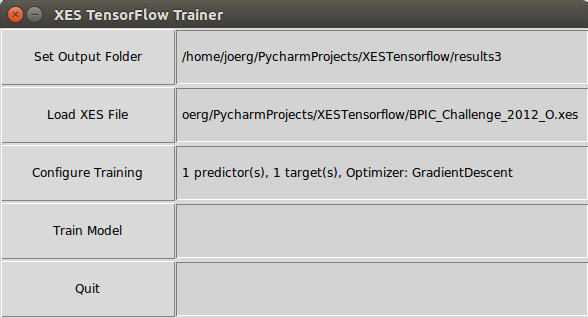}
\caption{Main screen of the training application}
\label{fig:main}
\end{figure}

Our software is implemented in Python 2.7 and uses Tensorflow 0.12 and Tkinter for the user interface. Figure~\ref{fig:main} shows the main screen that guides the user through the selection of an XES file, the configuration of the RNN and training parameters, to the training of the model.

\begin{figure}
\centering
\includegraphics[width=\textwidth]{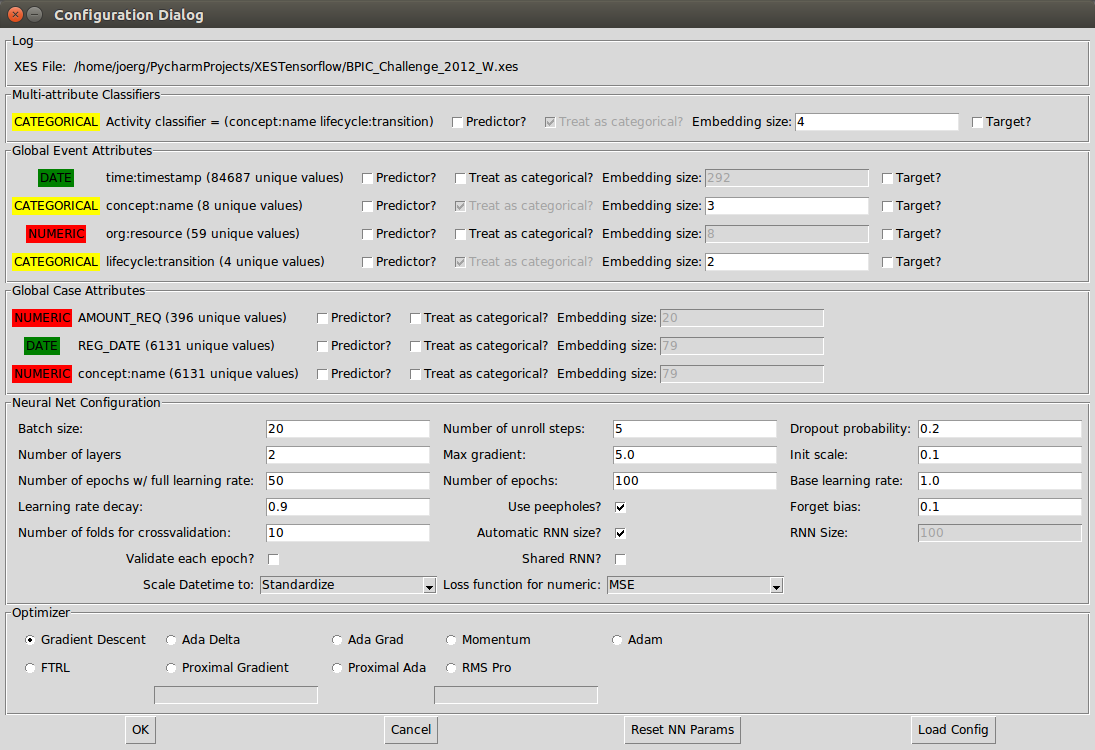}
\caption{Configuration screen. User can choose multiple predictors and targets, set general training parameters, and choose an optimizer.}
\label{fig:config}
\end{figure}

Figure~\ref{fig:config} shows the main configuration screen with sections for multi-attribute classifiers, global event and case attributes, RNN and training parameters and a choice of optimizer. Any classifier, global event and case attribute may be included as predictor, and classifiers and global event attributes may be chosen as predicted attributes (targets). The user can specify embedding dimensions for categorical attributes; the default values are the square root of the number of categories. The RNN configuration allows the user to specify batch size, number of unrolled steps, number of RNN layers, number of training epochs, etc. Users can specify an optional input mapping and the desired size of the RNN input vector. Finally, users have a choice of different gradient descent optimizers offered by Tensorflow and can adjust their hyperparameters. Configurations are automatically saved and the user can load saved configurations.

\begin{figure}
\centering
\includegraphics[scale=0.4]{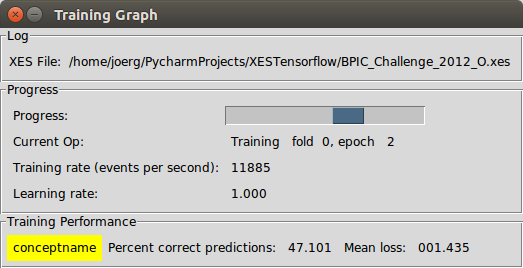}
\caption{Progress screen of the training process}
\label{fig:train}
\end{figure}

Figure~\ref{fig:train} shows the dialog indicating training progress. The current operation is shown, as well as the training rate (events/second) and the global learning rate for the gradient descent optimizer. For each predicted attribute, the latest training performance (proportion of correct prediction and loss function value) is shown.

%\vspace{-20pt}

Our focus is on providing a research tool for experimentation, rather than a production tool. Therefore, we have not made use of distributed Tensorflow and Tensorflow Serving. Tensorflow automatically allocates the compute operations to available CPUs and GPUs on a single machine. This provides adequate performance for the small size of typical event logs (megabyte rather than terabyte). We use our own prediction application with a simple API.

\section{Conclusion}

We presented a flexible deep-learning software application to predict business processes from industry-standard XES event logs. The software provides an easy-to-use graphical user interface for configuring predictors, targets, and parameters of the deep-learning prediction method. 

We have performed initial validation of the software to verify the correct operation of the software tool. Using the BPIC 2012 and 2013 datasets with the model and training parameters reported in \cite{EvermannRF2017}, this software tool replicates their training results in \cite{EvermannRF2017}. Current work with this software tool is ongoing, and focuses on using different combinations of predictors and targets, made possible by our flexible approach to handling predictors and constructing RNN inputs, to improve upon the state-of-the-art prediction performance.

\bibliographystyle{splncs03}

\end{document}